\documentclass[conference]{IEEEtran}
\IEEEoverridecommandlockouts
\usepackage{cite}
\usepackage{amsmath,amssymb,amsfonts}
\usepackage{algorithmic}

\usepackage{textcomp}
\usepackage{xcolor}
\usepackage[utf8]{inputenc} 
\usepackage[T1]{fontenc}    
\usepackage[hidelinks]{hyperref}       
\usepackage{url}            
\usepackage{booktabs}       
\usepackage{amsfonts}       
\usepackage{nicefrac}       
\usepackage{microtype}      
\usepackage{lipsum}
\usepackage{fancyhdr}       
\usepackage{graphicx,float}  

\usepackage{subcaption} 

\def\BibTeX{{\rm B\kern-.05em{\sc i\kern-.025em b}\kern-.08em
    T\kern-.1667em\lower.7ex\hbox{E}\kern-.125emX}}
\begin{document}

\title{Integrating Wearable Sensor Data and Self-reported Diaries for Personalized Affect Forecasting\vspace{15pt}}


\author{Zhongqi Yang, Yuning Wang, Ken S. Yamashita, Maryam Sabah, Elahe Khatibi,\\ Iman Azimi, Nikil Dutt, Jessica L. Borelli, and Amir M. Rahmani
}

\author{\IEEEauthorblockN{ Zhongqi Yang$^1$,
 Yuning Wang$^4$,
 Ken S. Yamashita$^2$,
Elahe Khatibi$^1$,
Iman Azimi$^1$,\\
Nikil Dutt$^1$,
Jessica L. Borelli$^2$,
and Amir M. Rahmani$^{1,3}$}
\IEEEauthorblockA{\textit{$^1$Department of Computer Science, University of California, Irvine}\\
\textit{$^2$Department of Psychological Science, University of California, Irvine}\\\textit{$^3$School of Nursing, University of California, Irvine}\\
\textit{$^4$Department of Computing, University of Turku}\\
\{zhongqy4, ksyamash, ekhatibi, azimii, dutt, jessica.borelli, a.rahmani\}@uci.edu, yuning.y.wang@utu.fi}}


\maketitle

\begin{abstract}
Emotional states, as indicators of affect, are pivotal to overall health, making their accurate prediction before onset crucial.
Current studies are primarily centered on immediate short-term affect detection using data from wearable and mobile devices.
These studies typically focus on objective sensory measures, often neglecting other forms of self-reported information like diaries and notes.
In this paper, we propose a multimodal deep learning model for affect status forecasting. 
This model combines a transformer encoder with a pre-trained language model, facilitating the integrated analysis of objective metrics and self-reported diaries.
To validate our model, we conduct a longitudinal study, enrolling college students and monitoring them over a year, to collect an extensive dataset including physiological, environmental, sleep, metabolic, and physical activity parameters, alongside open-ended textual diaries provided by the participants. Our results demonstrate that the proposed model achieves predictive accuracy of 82.50\% for positive affect and 82.76\% for negative affect, a full week in advance.
The effectiveness of our model is further elevated by its explainability. 
\end{abstract}


\section{Introduction}

Affect is a broad term referring to one's subjective feeling states. 
Affective states can influence moment-to-moment emotional experience as well as long-term mood and have impacts on overall well-being~\cite{ekici2014effect}.
Prolonged negative affect has been demonstrated to yield adverse consequences for both physical and mental health outcomes, including increased susceptibility to mental health conditions such as depression~\cite{frijda1987emotion,smidt2015brief,shankar2016effects,remes2021biological}, which ranks among the leading causes of disability~\cite{karrouri2021major}. Accurate identification, detection, and prediction of affect are crucial for effective intervention and prevention strategies.

In recent years, wearable and mobile devices, along with machine learning-based approaches, have been employed to track individuals' affective states~\cite{wang2022systematic}. 
Existing works focus on gathering and analyzing objective physiological and behavioral data through the use of laboratory-based or wearable/portable devices. 
Such research utilizing multimodal data, including electroencephalogram (EEG), photoplethysmogram (PPG), has primarily been conducted in laboratory settings~\cite{deb2017emotion, filippini2022automated, yang2020ai}. 
Expanding beyond the confines of laboratory environments, wearable and mobile devices such as smartwatches, smart rings, wristbands, and smartphones have been increasingly used for real-time, remote monitoring of individuals' affective states in free-living settings~\cite{kanjo2019deep,wang2024differential, wang2014studentlife, taylor2017personalized, jafarlou2023objective}.
Notably, \cite{jafarlou2023objective} demonstrates the feasibility of using fully objective measurements from commercial wearable devices to predict affect status in everyday settings.  
This study employs smart rings, smartwatches, and smartphones to continuously monitor physiological and environmental information, as well as sleep, metabolic, and physical activity patterns. 
Machine learning models are leveraged for data analysis to predict positive and negative affect statuses for the next day.
This research highlights that utilizing data collected from wearable devices enables a purely objective method to predict one's emotional state the following day. Although the accuracy of this approach is not exceptionally high, it demonstrates the potential of wearable technology in forecasting affect status.

However, there are two limitations of the existing studies. 
Firstly, while current remote affect monitoring systems primarily rely on objective data gathered from wearable devices in everyday settings, 
they
often ignore the valuable insights that can be obtained from human-generated textual data in daily life, which are crucial for mental health analysis. 
Such textual data play a pivotal role in providing information, such as life events or emotional narratives, which offer essential information for a comprehensive understanding of an individual's mood trajectory~\cite{suhasini2020emotion,baboo2022sentiment}. 
Incorporating such textual data (including diaries, notes, etc.) complementing the objective measurements still requires further exploration.
Secondly, existing research concentrates on the instantaneous detection or short-term prediction of affective states.
The reliance on fully collected data from the next day limits their predictive capabilities to no earlier than the following day.
This timeframe may not be adequate for timely prevention strategies.
There is a pressing need for research focused on forecasting mental health conditions significantly in advance of their onset to enhance preventive measures and overall mental well-being.
 
In this paper, we present a multimodal deep learning model which is based on a transformer encoder and a pre-trained language model, DistilBERT, to forecast affect status one week in advance.
We accomplish this by fusing objective features from wearable and mobile devices and self-reported diaries. 
The objective features encompass a range of data, including sleep patterns, physiological metrics, physical activity, metabolic rates, and environmental factors.
For the diary data, we extract features from two primary aspects: the diary submission frequency and the diary content. 
To evaluate the effectiveness of our proposed affect forecasting model, we collect a dataset featuring multimodal information from wearable devices and self-reported diaries detailing daily affective highs and lows.
This system also includes 20 unique discrete affect ratings provided by participants daily and weekly.
The affect ratings are aggregated to establish labels for positive and negative affect statuses.
Furthermore, we investigate model explainability by analyzing the contribution of features as indicated by Shapley values, and the attention scores assigned to keywords in the diaries.

\section{Data Collection}\label{sec:data}

We collected data for assessing mental health and affect status of college students including 25 undergraduate students between the ages of 18 and 22 who could speak and write English fluently. 
Participants could not be parents or married or returning to school after a 3 year period or longer to maintain a homogeneous sample that reflects the college student population. 
Participants must  not have participated in any of 
our previous studies on this topic.

Over the course of the study, we tracked participants’ emotional states, physiological patterns, and behavioral habits through smart devices and ecological momentary assessments (EMAs).
Participants were fitted with the Oura ring and Samsung Gear Sport smartwatch and downloaded the corresponding Oura and Samsung Android mobile apps in an effort to capture an accurate depiction of their daily physical habits, sleep, and health. These devices were integrated into the ZotCare mHealth platform \cite{zotcare}. 
The Oura ring collects data on sleep and cardiovascular activity that are used in this study to assess sleep quality by measuring sleep duration, average heart rate during sleep, and heart rate variability during sleep. 
The Samsung watch complements the physical activity feature set by adding walk and run steps and captures the atmospheric pressure, which enriches the environmental features available for analysis.
The detailed list of the objective features is the same as those described in~\cite{jafarlou2023objective}.

Participants utilized the ZotCare application to submit self-reports functioning as daily diaries through 
EMAs.
These entries vary from expressions of personal sentiments and moods to narratives of daily experiences. 
Participants document key moments of their day, detailing the high and low points of their day's affective experiences, overall emotional experiences, and notable daily events such as physical activities, academic exams, or social interactions.

Participants were also asked to engage with the ZotCare app to submit their daily and weekly affect status through EMAs that were used as labels with a one-week delay.
The affect status is evaluated using a scale ranging from 0 (``Very Slightly") to 100 (``Extremely"),
, where 0 meant ``Very Slightly" and 100 ``Extremely", 
including 20 affect words like ``inspired" and ``nervous," based on the Positive and Negative Affect Schedule (PANAS)~\cite{sano2016measuring,watson1988development}.
Each affect word is scored individually and as part of a composite, with Positive Affect (PA) and Negative Affect (NA) calculated as the average of 10 positive and negative items, respectively. 
Binary classifications for PA and NA are assigned based on whether an emotion's score surpassed or did not meet the median for all participants, thereby maintaining an equitable binary label distribution.
In line with standard practice, we exclude the central 20\% of values from the overall distribution prior to assigning labels.

\section{Affect Forecasting Model}
\begin{figure}[!t]
\vspace{-5mm}
    \centering
    \includegraphics[width=0.5\textwidth]{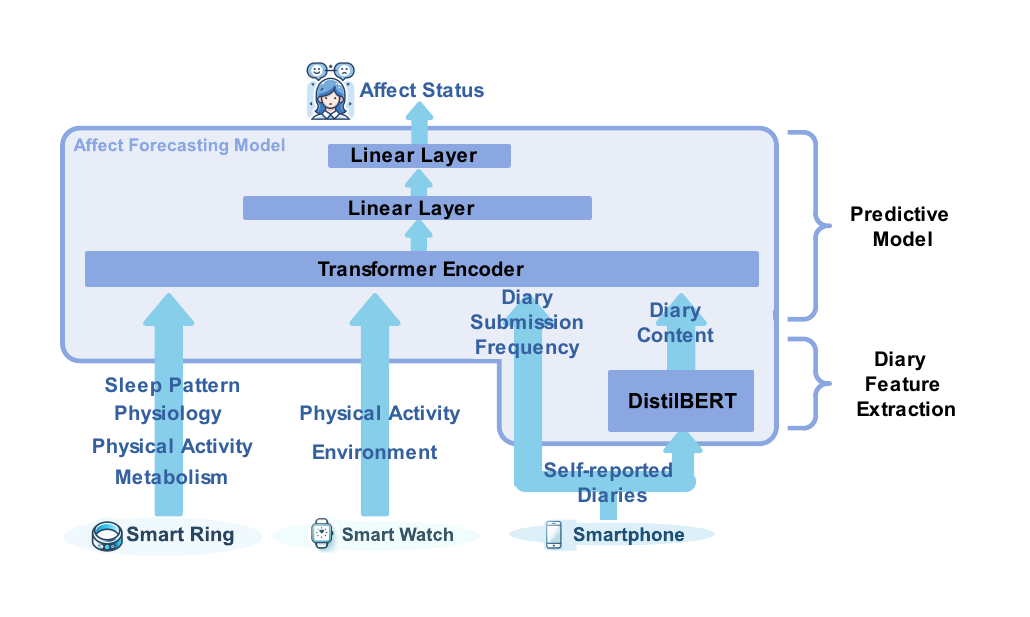}
    \vspace{-8mm}
    \caption{Affect Forecasting Model Workflow.}

    \label{fig:workflow}
     \vspace{-5mm}
\end{figure}

This section introduces the affect forecasting model designed for the analysis of our multimodal data.
The workflow of the model is illustrated in Figure~\ref{fig:workflow}.

\subsection{Diary Feature Extraction}\label{sec:textfeature}

Participants submitted daily self-reported diaries to describe their daily feelings through the ZotCare application on smartphones.
We introduce the self-reported diary feature extraction of the affect forecasting model for two key features from the diaries: Diary Content and Diary Submission Frequency.

\paragraph{Diary Content}

The analysis of self-reported diaries poses a unique challenge in affective computing as they usually do not explicitly reveal emotional status.
Moreover, these diaries rarely directly indicate physiological or behavioral factors that could predict future affect status. 
Considering the model efficiency, we employ DistilBERT~\cite{sanh2019distilbert} complemented by a linear layer to extract meaningful insights from the diaries. 
DistilBERT functions to parse and interpret the underlying sentiments of the diaries and offers a high-dimensional vector representation of the input.
Then the following linear layer acts on the high-dimensional representation produced by DistilBERT, transforming them into a streamlined, singular value vector as the Diary Content feature. 

\paragraph{Diary Submission Frequency}

Another challenge in our study is the variability in the frequency of diary submissions by participants, serving as a factor that, as highlighted in recent studies, reflects the mental health status of individuals~\cite{vishnubhotla2022tweet}. 
We calculate the diary submission frequency based on how many days a participant submits in a one-week window.

\subsection{Predictive Model}

The architecture of the predictive model is built on a Transformer Encoder~\cite{vaswani2017attention} to aggregate the features from a wide range of modalities efficiently through its self-attention mechanism.
Following the transformer encoder, we integrate a 2-layer multi-layer Perceptron to project the complex embeddings generated by the transformer into a singular, interpretable value that directly corresponds to the affect status. 

The input for the predictive model amalgamates diary features derived from Self-reported Diary Processing with objective features, as detailed in Section~\ref{sec:data}.
To achieve this, we concatenate both objective features and diary features to create an extensive feature vector.

\subsection{Model Training Strategy}
\begin{figure}[!b]
\vspace{-9mm}
    \centering

    \includegraphics[width=0.4\textwidth]{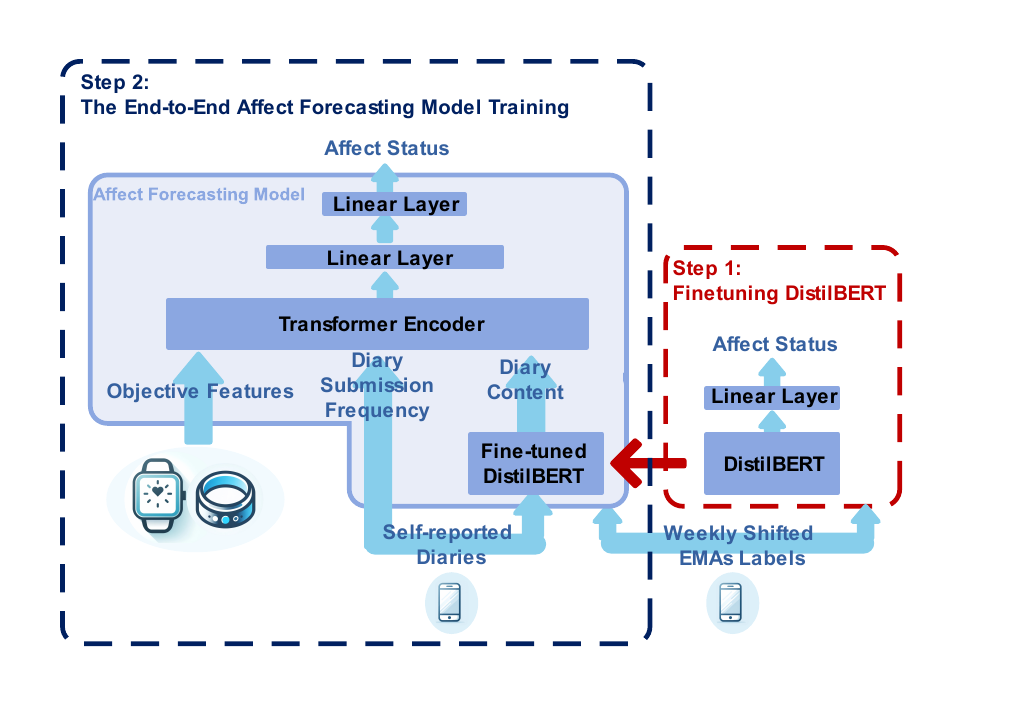}
         \vspace{-4mm}
    \caption{Training Process of the Affect Forecasting Model}
    \label{fig:training}
   
\end{figure}

The integration of diary features in our predictive model is challenged by the initial ineffectiveness of DistilBERT.
To address this, we propose a sequential training process that is executed in two steps.
Initially, DistilBERT is fine-tuned independently to adapt its pre-trained weights to our specific affective context. 
We then train the entire model in a joint manner. 
This sequential approach ensures a balanced learning process to prevent any type of feature from disproportionately influencing the model's development.
This process is also illustrated in Figure~\ref{fig:training}.

\paragraph{Step 1: Fine-tuning DistilBERT}

We observe that it is challenging when training 
DistilBERT and the predictive model jointly.
Given that DistilBERT is a general-purpose model trained on a diverse corpus, 
we must
align DistilBERT to the affect forecasting task. 
However, jointly training DistilBERT with the predictive model presents challenges. 
Initially, DistilBERT's performance in identifying affect-related information from diary entries is suboptimal, leading the predictive model to assign attention lower scores to the diary features. 
This leads to weak gradient signals being sent back to DistilBERT during training 
that may result in omission of the diary information. 

To address this challenge, it is essential to customize DistilBERT specifically for the affect forecasting task. Our approach involves initially fine-tuning DistilBERT to better suit affect forecasting. We accomplish this by integrating a temporary linear layer on top of DistilBERT, which serves to translate embeddings into predicted affect status and facilitates gradient propagation. 
Both DistilBERT and this linear layer are then fine-tuned together on the training dataset.

\paragraph{Step 2: The End-to-End Affect Forecasting Model Training}

Following the initial fine-tuning of DistilBERT, Step 2 
jointly trains the whole affect forecasting model encompassing the fine-tuned DistilBERT, the transformer encoder, and the linear layers.
Diaries are processed through DistilBERT to extract diary content feature, which are subsequently concatenated with objective features and diary submission frequency. 
The combined inputs are then fed through the transformer encoder, followed by linear layers that map these inputs to the corresponding affect status.

\subsection{Model Explainability}
We demonstrate the explainability of the proposed model in two aspects.
We first highlight how individual features impact the forecasted affect status.
We accomplish this by employing path-dependent feature perturbation algorithms provided by the SHAP (SHapley Additive exPlanations) library~\cite{lundberg2017unified} to calculate the Shapley values for each feature. 
Additionally, to gain a deeper understanding of which keywords the model identifies as crucial for predicting affect status, we delve into the significance of specific words in the diary entries. 
This is achieved by analyzing the attention scores allocated by the fine-tuned DistilBERT model to these keywords.

\section{Results}

\begin{table}[b!]
\vspace{-5mm}
    \caption{Forecasting Accuracy Results}
    \centering
    \begin{tabular}{|c|c|c|}
    \hline
    & Personalized (\%) & Non-personalized (\%) \\
    \hline
       Forecasting PA w/ Diary&  \textbf{82.50} &81.20 \\
       \hline
       Forecasting NA w/ Diary&\textbf{87.76} &80.22\\
        \hline
        Forecasting PA w/o Diary& 81.08 &79.03\\
        \hline
        Forecasting NA w/o Diary&82.11 &75.31\\
        \hline

    \end{tabular}
    \label{tab:texteffect}
\end{table}


\begin{figure}[]
    \centering
    
    \includegraphics[width=0.40\textwidth]{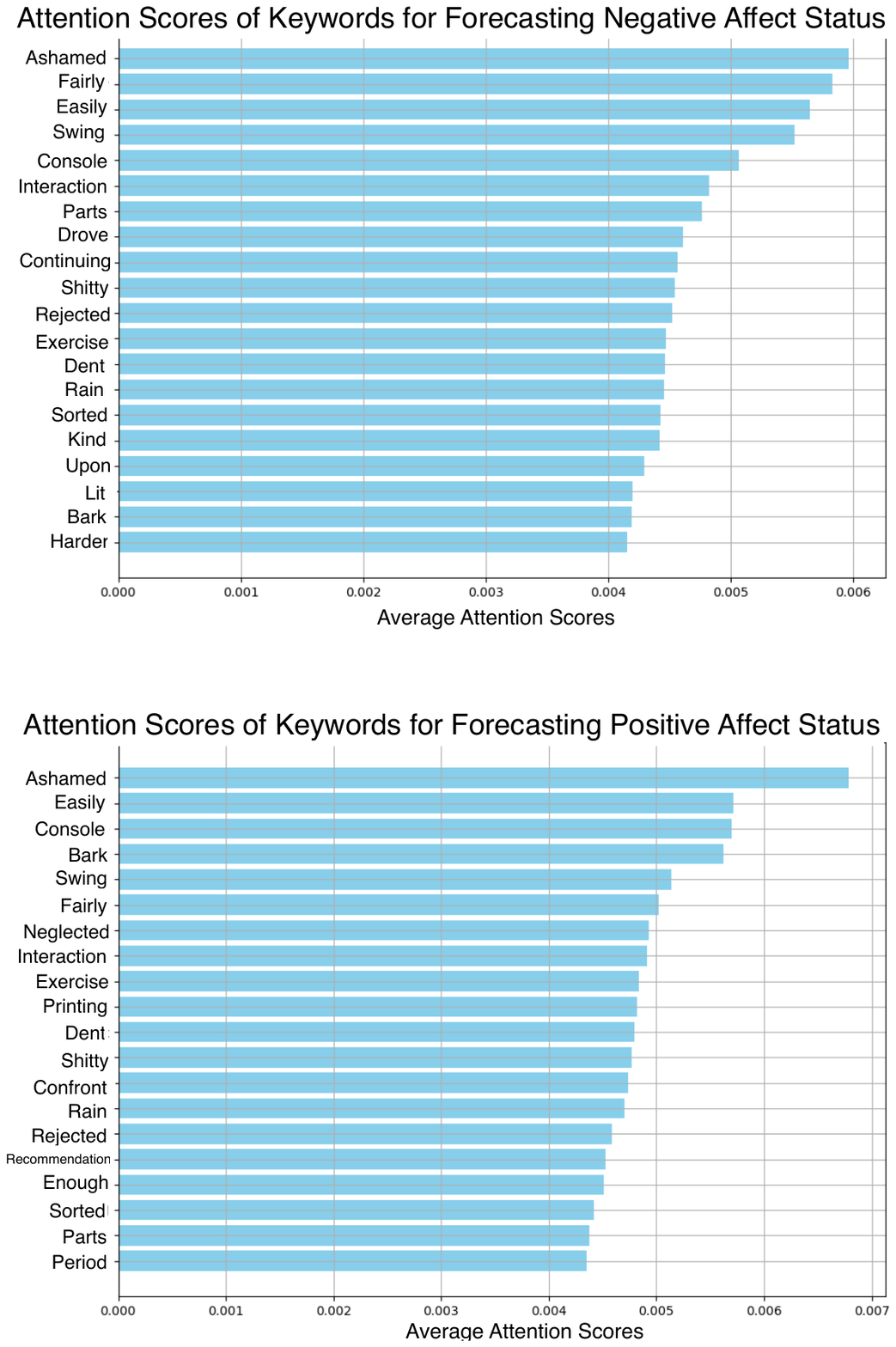}
    \caption{Attention scores of keywords}
    \label{fig:attention}
     \vspace{-4mm}
\end{figure}
\begin{figure}[htb]
    \centering
    \vspace{-10mm}
    \includegraphics[width=0.38\textwidth]{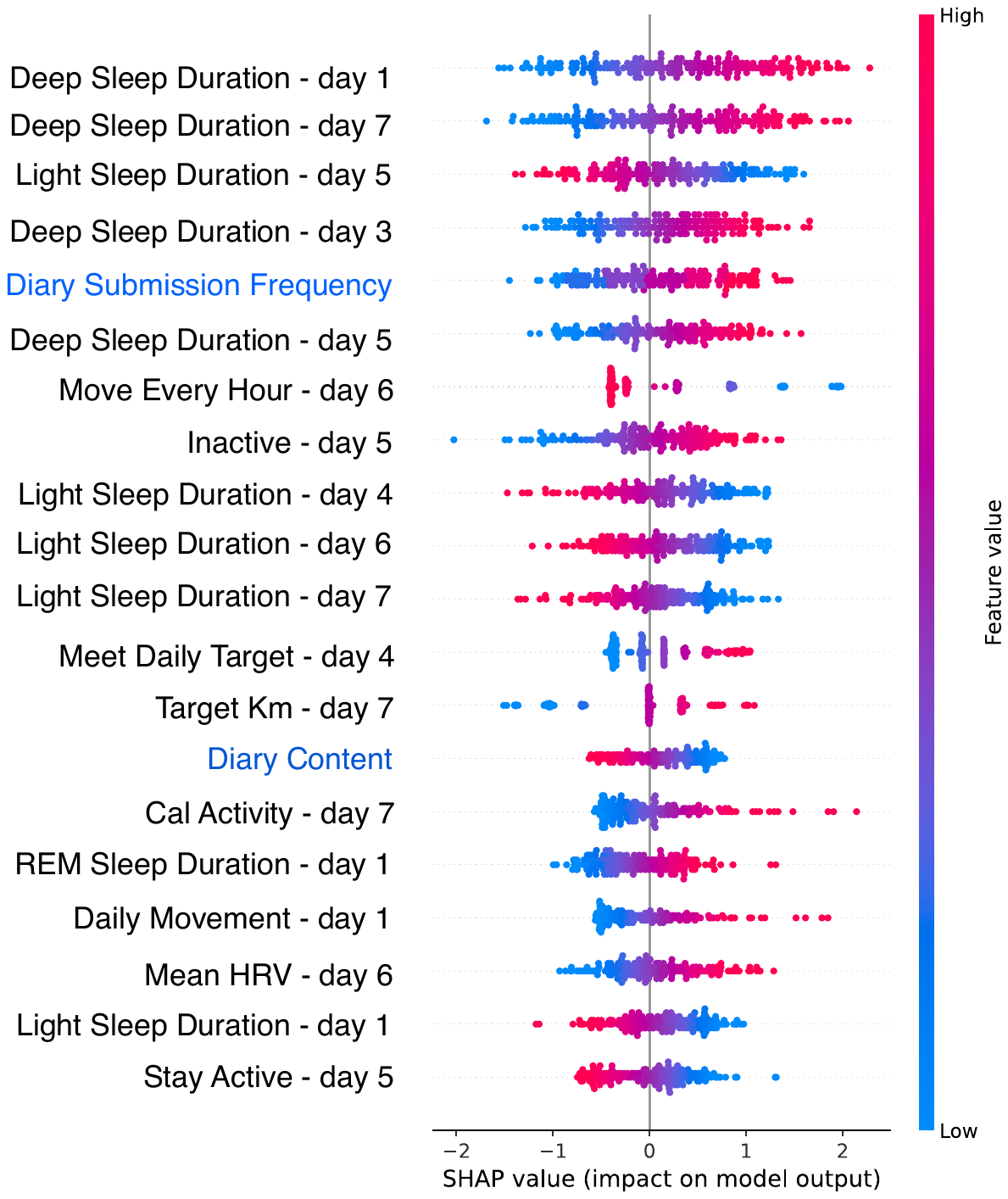}
    \vspace{-10mm}
    \caption{Feature Shapley Values for Positive Affect Forecasting}
    \label{fig:shappos}
     \vspace{-6mm}
\end{figure}

\begin{figure}[htb]
\vspace{-10mm}
    \centering
      
    \includegraphics[width=0.38\textwidth]{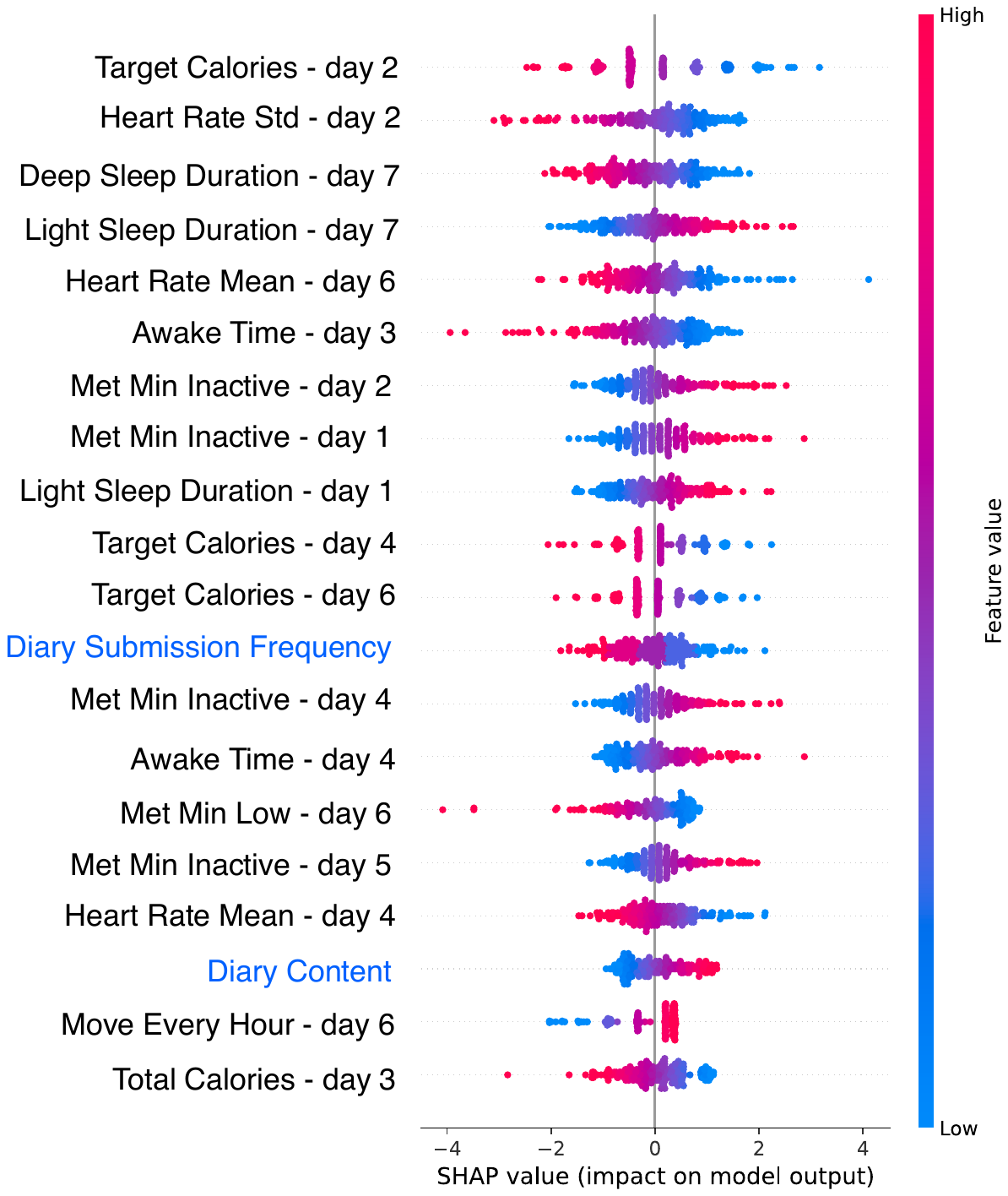}
    \vspace{-11mm}
    \caption{Feature Shapley Values for Negative Affect Forecasting}
    \label{fig:shapneg}
   \vspace{-6mm}
\end{figure}

We evaluate the proposed model using leave-one-subject-out cross-validation and a personalized cross-validation strategy. 
In the leave-one-subject-out cross-validation approach, each iteration involves training the model on the data from all but one subject and testing on the data from the excluded subject.
For the personalized cross-validation, we utilize the latter half of a participant’s data for testing, while the initial half, combined with data from other participants, is employed for training. 
This method ensures that each iteration of the model incorporates historical data from an individual participant for training, fostering a personalized model. 
We refer to the models derived from personalized cross-validation as 'Personalized,' while those obtained from leave-one-subject-out cross-validation are designated as 'Non-personalized.'
The accuracy of our proposed model for forecasting affect status are shown in Table~\ref{tab:texteffect}.

We observe an enhancement in accuracy in the personalized cross-validation process that the personalized models, distinct for the training on the participant's unique data set, outperform generic, non-personalized models.
This result suggests that models tailored to individual behavioral and emotional patterns yield more precise and reliable predictions, emphasizing the value of personalization in predictive analytics.

Additionally, to illustrate the impact of the diary features, we compare the proposed model against a purely objective model, which excluded the diaries and relies solely on objective features, designated as 'Forecasting w/o Diary'.
Our findings indicate that the inclusion of diary features leads to a boost in the model's forecasting accuracy.

\subsection{Features' Shapley Value}

The Shapley values distribution of features are shown in Figure~\ref{fig:shappos} and Figure~\ref{fig:shapneg}.
We observe that both light and deep sleep duration emerge as top-ranked predictors, with higher duration of deep sleep and lower light sleep duration contributing strongly to the prediction of a positive affect status, and lower duration is correlated to negative affect status.
Diary Submission Frequency also holds considerable weight as the higher frequency of submitted diaries indicates an association with positive affect status while a lower frequency correlates with an increase in negative affect status.
The physically active time and the total movement contribute to our model's positive affect predictions.
Moreover, we observe that dietary metrics, such as calorie intake, along with activity metrics like MET and daily target calorie expansion, are influential in determining negative affect status, underscoring the multifaceted nature of affect status predictions.



\subsection{Diary Keywords Attention Scores}

Recognizing the critical role of diary content in forecasting, we 
analyze the most influential keywords identified by their attention scores computed by DistillBERT.
Figures~\ref{fig:attention} demonstrate the attention scores derived from the DistilBERT model for forecasting both positive and negative affect status. 
The most influential keywords are largely similar across both positive and negative affect forecasting.
It can be seen that the words demonstrating emotional status, such as "ashamed", "easily", and "shitty" obtain higher attention scores.
Additionally, keywords pertaining to specific life events or activities, like "drove", "console", "exercise", and "rain", are also identified as significant in affect forecasting.



\section{Discussion}

The results of this study not only validate the feasibility of one-week forecasting but also align with the accuracy of next-day predictions reported in our previous study~\cite{jafarlou2023objective}, highlighting the effectiveness of our multimodal approach, especially incorporating self-reported diaries.
Although forecasting a single loneliness level was previously shown to be feasible using objective features~\cite{yang2023loneliness}, affect status forecasting, as a multi-facet case, is still challenging.
We show that the integration of diary data with objective modalities, such as physiological, environmental, and physical activity information, gathered from wearable and mobile devices represents a novel and efficient approach to affective computing. 


In terms of explainability, we find that levels of sleep had the greatest impact on the model output. 
This is consistent with previous literature relating the quality of sleep to mood, where sleep deprivation is found to have a strong influence on the positive affective system \cite{finan2017partial}. 
Our findings also highlight certain keywords with higher attention scores in affect forecasting.
The significant terms are similar across both positive and negative affect. 
Words demonstrating emotional status (i.e., ashamed) have a higher attention score in general, which is consistent with previous research suggesting that emotionally intense words can influence behavior and cognition \cite{carretie2008modulation}.
Additionally, studies have found that negative words are generally stronger in extremity in comparison to positive words \cite{yang2013positive}. 
This aligns with our findings, as the highest score of emotion-tied terms has a negative valence. We find keywords relating to day-to-day activities (i.e., rain, console) are significant in affect forecasting. 
Additionally, this study indicates that a higher frequency of diary submissions is associated with an increased likelihood of positive affect status and a decreased likelihood of negative affect status.
Some research has found links between writing and mood fluctuations; for instance, a study found that completing daily diary entries reduced self-reported depression levels as well as influenced emotion regulation strategies~\cite{suhr2017maintaining}.
Although neutral terminology may not seem significant, simply writing and sharing daily activities may have implications on mood \cite{yau2021coping}. 
Future work could expand on specific keyword frequency, studying if the increased presence of a specific valence (positive or negative) can be associated with positive affect or negative affect forecasting, respectively.

\section{Conclusion}

This study introduced a multimodal deep learning model to predict affect status using a combination of data from wearable devices and self-reported diaries. The model exhibited satisfactory accuracy of 82.50\% and 87.76\% for positive and negative affect, respectively, affirming the practicality of forecasting affect status.
A significant outcome of our study is the improvement in accuracy upon incorporating diary data, underscoring the contribution of self-reported diaries to this process.
Moreover, our results stressed the importance of personalized methods in monitoring mental health, suggesting that personalized approaches yield more accurate outcomes.
Future research could expand upon this study by applying the model to diverse populations and environments. This would aim to enhance the integration of both objective and self-reported diary data for more accurate predictions. 
\bibliographystyle{unsrt}
\bibliography{references}

\begin{thebibliography}{10}

\bibitem{ekici2014effect}
Berkay Ekici, Ebru~Akgul Ercan, Sengul Cehreli, and Hasan~Fehmi T{\"o}re.
\newblock The effect of emotional status and health-related quality of life on the severity of coronary artery disease.
\newblock {\em Kardiologia Polska (Polish Heart Journal)}, 72(7):617--623, 2014.

\bibitem{frijda1987emotion}
Nico~H Frijda.
\newblock Emotion, cognitive structure, and action tendency.
\newblock {\em Cognition and emotion}, 1(2):115--143, 1987.

\bibitem{smidt2015brief}
Katharine~E Smidt and Michael~K Suvak.
\newblock A brief, but nuanced, review of emotional granularity and emotion differentiation research.
\newblock {\em Current Opinion in Psychology}, 3:48--51, 2015.

\bibitem{shankar2016effects}
Nilani~L Shankar and Crystal~L Park.
\newblock Effects of stress on students' physical and mental health and academic success.
\newblock {\em International Journal of School \& Educational Psychology}, 4(1):5--9, 2016.

\bibitem{remes2021biological}
Olivia Remes, Jo{\~a}o~Francisco Mendes, and Peter Templeton.
\newblock Biological, psychological, and social determinants of depression: a review of recent literature.
\newblock {\em Brain sciences}, 11(12):1633, 2021.

\bibitem{karrouri2021major}
Rabie Karrouri, Zakaria Hammani, Roukaya Benjelloun, and Yassine Otheman.
\newblock Major depressive disorder: Validated treatments and future challenges.
\newblock {\em World journal of clinical cases}, 9(31):9350, 2021.

\bibitem{wang2022systematic}
Yan Wang, Wei Song, Wei Tao, Antonio Liotta, Dawei Yang, Xinlei Li, Shuyong Gao, Yixuan Sun, Weifeng Ge, Wei Zhang, et~al.
\newblock A systematic review on affective computing: Emotion models, databases, and recent advances.
\newblock {\em Information Fusion}, 83:19--52, 2022.

\bibitem{deb2017emotion}
S.~Deb et~al.
\newblock Emotion classification using segmentation of vowel-like and non-vowel-like regions.
\newblock {\em IEEE Trans Affect Comput.}, 10(3):360--73, 2017.

\bibitem{filippini2022automated}
Chiara Filippini, Adolfo Di~Crosta, Rocco Palumbo, David Perpetuini, Daniela Cardone, Irene Ceccato, Alberto Di~Domenico, and Arcangelo Merla.
\newblock Automated affective computing based on bio-signals analysis and deep learning approach.
\newblock {\em Sensors}, 22(5):1789, 2022.

\bibitem{yang2020ai}
Cheng-Jie Yang, Nicolas Fahier, Chang-Yuan He, Wei-Chih Li, and Wai-Chi Fang.
\newblock An ai-edge platform with multimodal wearable physiological signals monitoring sensors for affective computing applications.
\newblock In {\em 2020 IEEE International Symposium on Circuits and Systems (ISCAS)}, pages 1--5. IEEE, 2020.

\bibitem{kanjo2019deep}
Eiman Kanjo, Eman~MG Younis, and Chee~Siang Ang.
\newblock Deep learning analysis of mobile physiological, environmental and location sensor data for emotion detection.
\newblock {\em Information Fusion}, 49:46--56, 2019.

\bibitem{wang2024differential}
Ziyu Wang, Zhongqi Yang, Iman Azimi, and Amir~M Rahmani.
\newblock Differential private federated transfer learning for mental health monitoring in everyday settings: A case study on stress detection.
\newblock {\em arXiv preprint arXiv:2402.10862}, 2024.

\bibitem{wang2014studentlife}
Rui Wang, Fanglin Chen, Zhenyu Chen, Tianxing Li, Gabriella Harari, Stefanie Tignor, Xia Zhou, Dror Ben-Zeev, and Andrew~T Campbell.
\newblock Studentlife: assessing mental health, academic performance and behavioral trends of college students using smartphones.
\newblock In {\em Proceedings of the 2014 ACM international joint conference on pervasive and ubiquitous computing}, pages 3--14, 2014.

\bibitem{taylor2017personalized}
Sara Taylor, Natasha Jaques, Ehimwenma Nosakhare, Akane Sano, and Rosalind Picard.
\newblock Personalized multitask learning for predicting tomorrow's mood, stress, and health.
\newblock {\em IEEE Transactions on Affective Computing}, 11(2):200--213, 2017.

\bibitem{jafarlou2023objective}
Salar Jafarlou, Jocelyn Lai, Iman Azimi, Zahra Mousavi, Sina Labbaf, Ramesh~C Jain, Nikil Dutt, Jessica~L Borelli, Amir Rahmani, et~al.
\newblock Objective prediction of next-day’s affect using multimodal physiological and behavioral data: Algorithm development and validation study.
\newblock {\em JMIR Formative Research}, 7(1):e39425, 2023.

\bibitem{suhasini2020emotion}
Matla Suhasini and Badugu Srinivasu.
\newblock Emotion detection framework for twitter data using supervised classifiers.
\newblock In {\em Data Engineering and Communication Technology: Proceedings of 3rd ICDECT-2K19}, pages 565--576. Springer, 2020.

\bibitem{baboo2022sentiment}
S~Santhosh Baboo and M~Amirthapriya.
\newblock Sentiment analysis and automatic emotion detection analysis of twitter using machine learning classifiers.
\newblock {\em International Journal of Mechanical Engineering}, 7(2), 2022.

\bibitem{zotcare}
Sina Labbaf, Mahyar Abbasian, Iman Azimi, Nikil Dutt, and Amir~M. Rahmani.
\newblock Zotcare: a flexible, personalizable, and affordable mhealth service provider.
\newblock {\em Frontiers in Digital Health}, 5, 2023.

\bibitem{sano2016measuring}
Akane Sano.
\newblock {\em Measuring college students' sleep, stress, mental health and wellbeing with wearable sensors and mobile phones}.
\newblock PhD thesis, Massachusetts Institute of Technology, 2016.

\bibitem{watson1988development}
David Watson, Lee~Anna Clark, and Auke Tellegen.
\newblock Development and validation of brief measures of positive and negative affect: the panas scales.
\newblock {\em Journal of personality and social psychology}, 54(6):1063, 1988.

\bibitem{sanh2019distilbert}
Victor Sanh, Lysandre Debut, Julien Chaumond, and Thomas Wolf.
\newblock Distilbert, a distilled version of bert: smaller, faster, cheaper and lighter.
\newblock {\em arXiv preprint arXiv:1910.01108}, 2019.

\bibitem{vishnubhotla2022tweet}
Krishnapriya Vishnubhotla and Saif~M Mohammad.
\newblock Tweet emotion dynamics: Emotion word usage in tweets from us and canada.
\newblock {\em arXiv preprint arXiv:2204.04862}, 2022.

\bibitem{vaswani2017attention}
Ashish Vaswani, Noam Shazeer, Niki Parmar, Jakob Uszkoreit, Llion Jones, Aidan~N Gomez, {\L}ukasz Kaiser, and Illia Polosukhin.
\newblock Attention is all you need.
\newblock {\em Advances in neural information processing systems}, 30, 2017.

\bibitem{lundberg2017unified}
Scott~M Lundberg et~al.
\newblock A unified approach to interpreting model predictions.
\newblock {\em NIPS'17}, 30, 2017.

\bibitem{yang2023loneliness}
Zhongqi Yang, Iman Azimi, Salar Jafarlou, Sina Labbaf, Jessica Borelli, Nikil Dutt, and Amir Rahmani.
\newblock Loneliness forecasting using multi-modal wearable and mobile sensing in everyday settings.
\newblock {\em medRxiv}, pages 2023--06, 2023.

\bibitem{finan2017partial}
Patrick~H Finan, Phillip~J Quartana, Bethany Remeniuk, Eric~L Garland, Jamie~L Rhudy, Matthew Hand, Michael~R Irwin, and Michael~T Smith.
\newblock Partial sleep deprivation attenuates the positive affective system: effects across multiple measurement modalities.
\newblock {\em Sleep}, 40(1):zsw017, 2017.

\bibitem{carretie2008modulation}
Luis Carreti{\'e}, Jos{\'e}~A Hinojosa, Jacobo Albert, Sara L{\'o}pez-Mart{\'\i}n, Bel{\'e}n~S De~La~G{\'a}ndara, Jos{\'e}~M Igoa, and Mar{\'\i}a Sotillo.
\newblock Modulation of ongoing cognitive processes by emotionally intense words.
\newblock {\em Psychophysiology}, 45(2):188--196, 2008.

\bibitem{yang2013positive}
Jiemin Yang, Jing Zeng, Xianxin Meng, Liping Zhu, Jiajin Yuan, Hong Li, and Nasir Yusoff.
\newblock Positive words or negative words: Whose valence strength are we more sensitive to?
\newblock {\em Brain research}, 1533:91--104, 2013.

\bibitem{suhr2017maintaining}
M~Suhr, AK~Risch, and G~Wilz.
\newblock Maintaining mental health through positive writing: Effects of a resource diary on depression and emotion regulation.
\newblock {\em Journal of clinical psychology}, 73(12):1586--1598, 2017.

\bibitem{yau2021coping}
Joanna~C Yau, Stephanie~M Reich, and Tao-Yi Lee.
\newblock Coping with stress through texting: an experimental study.
\newblock {\em Journal of Adolescent Health}, 68(3):565--571, 2021.

\end{thebibliography}

\end{document}